\documentclass{article}

    \usepackage[preprint, nonatbib]{neurips_2023}

\usepackage[utf8]{inputenc} 
\usepackage[T1]{fontenc}    
\usepackage[colorlinks=true, linkcolor=blue]{hyperref}       
\usepackage{url}            
\usepackage{booktabs}       
\usepackage{amsfonts}       
\usepackage{nicefrac}       
\usepackage{microtype}      
\usepackage{xcolor}         
\usepackage{makecell}

\usepackage{graphicx}
\usepackage{wrapfig}

\usepackage{amsmath}
\usepackage{amssymb}
\usepackage{environ}
\usepackage{slashed}
\usepackage{mathrsfs}
\usepackage{subcaption}

\usepackage{algorithm}
\usepackage{algpseudocode}

\usepackage{footnote}

\usepackage{multirow}
\usepackage{array, booktabs}
\newcolumntype{C}[1]{>{\centering\let\newline\\\arraybackslash\hspace{0pt}}m{#1}}
\newcolumntype{M}[1]{>{\centering\arraybackslash}m{#1}}

\NewEnviron{derivation}
{%
    \\
    \begin{minipage}{\linewidth}
        \begin{equation}\begin{split}
        \BODY
        \end{split}\end{equation}
    \end{minipage}
}

\makeatletter
\newcommand{\mfootnote}[1]{%
  \ifmeasuring@
    \chardef\@tempfn=\value{footnote}%
    \! \footnotemark
    \setcounter{footnote}{\@tempfn}%
  \else
    \iffirstchoice@
      \! \footnote{#1}%
    \fi
  \fi}
\makeatother

\title{QUICK: Quantization-aware Interleaving and Conflict-free Kernel for efficient LLM inference}

\author{
\makecell{  Taesu Kim \hspace{15pt}
            Jongho Lee \hspace{15pt}
            Daehyun Ahn \hspace{15pt}
            Sarang Kim \\
            Jiwoong Choi \hspace{15pt}
            Minkyu Kim \hspace{15pt}
            Hyungjun Kim}\\
            \\
SqueezeBits Inc.
}

\begin{document}

\graphicspath{ {figures/} }

\maketitle

\begin{abstract}
We introduce QUICK, a group of novel optimized CUDA kernels for the efficient inference of quantized Large Language Models (LLMs). QUICK addresses the shared memory bank-conflict problem of state-of-the-art mixed precision matrix multiplication kernels. Our method interleaves the quantized weight matrices of LLMs offline to skip the shared memory write-back after the dequantization. We demonstrate up to 1.91x speedup over existing kernels of AutoAWQ on larger batches and up to 1.94x throughput gain on representative LLM models on various NVIDIA GPU devices.

\textbf{Code: }\hyperlink{https://github.com/SqueezeBits/QUICK}{https://github.com/SqueezeBits/QUICK}
\end{abstract}

\section{Introduction}
Enhancing the efficiency of Large Language Models (LLMs) has become increasingly crucial due to the escalating demand for deploying state-of-the-art models in real-world scenarios \cite{brown2020gpt3,jiang2023mistral,jiang2024mixtral,roumeliotis2023llamav2,touvron2023llama}. The improved performance of LLMs is attributed to their growing size, characterized by a trend toward larger models with parameter counts exceeding several hundred billion. However, the substantial size of these models has necessitated the adoption of model compression techniques such as quantization and pruning \cite{ashkboos2024slicegpt,dettmers2022llmint8,frantar2023sparsegpt,lee2024owq,lin2023awq,xiao2023smoothquant}.

Among these techniques, weight-only quantization has garnered significant attention for its potential to compress the memory footprint of LLMs \cite{frantar2022gptq,lee2024owq,lin2023awq}. This approach aims to reduce model size and accelerate computation by quantizing weights to smaller bit sizes while retaining activation tensors at higher precision. Consequently, there is a growing need for fast mixed-precision General Matrix Multiplication (GEMM) kernels to support such operations.

Despite these advancements, existing open-source kernels for mixed-precision GEMM have demonstrated limitations in throughput, primarily due to the overhead associated with weight dequantization. Analysis of these kernels has revealed shared memory write-back bank conflicts during the dequantization process as a significant bottleneck. Leveraging this insight, we introduce QUICK, a solution designed to mitigate shared memory bank conflicts by reordering weight matrices offline.

\section{Preliminary}
\subsection{Quantization and Dequantization}
Quantization involves the reduction of precision or range of a continuous variable to a discrete set of values. This process is commonly employed to decrease the bit precision of tensors, thereby reducing the memory footprint of Neural Network models. When supported by appropriate computation kernels, quantization enables acceleration of the models with low-precision computations.

Given that LLMs typically encompass billions of parameters, researchers have explored quantization as a means to reduce memory usage and improve inference efficiency. Specifically, weight-only quantization focuses solely on quantizing the weights of the model while maintaining activations at a higher precision, such as 16-bit floating point \cite{frantar2022gptq,lee2024owq,lin2023awq}. This strategy effectively reduces memory requirements by representing weights with fewer bits while retaining activation tensors in floating point precision.

Weight-only quantization is generally recognized for dramatically reducing the memory requirements and preserving the performance of LLMs. However, since activations remain in higher precision, weights must undergo dequantization back to higher precision before being multiplied by activations during inference. This dequantization process has minimal impact on inference efficiency when the batch size is 1 since the computation is mainly memory-bounded in such case. However, for larger batch sizes, GEMMs are mostly computation-bounded, where mixed-precision GEMM operations become slower than their floating-point counterparts due to the overhead associated with dequantization.

\subsection{GEMM kernel using Tensor Core}
A substantial portion of the computational workload associated with LLMs primarily comprises GEMMs. Optimizing GEMM operations plays a pivotal role in enhancing the overall efficiency of LLM inference. Particularly on NVIDIA GPUs, GEMM computation has relied on the tiling strategy, which is widely employed to maximize memory reuse through the utilization of shared memory, thereby achieving a more favorable compute-memory ratio.

\begin{figure}
    \centering
    \includegraphics[width=0.7\columnwidth]{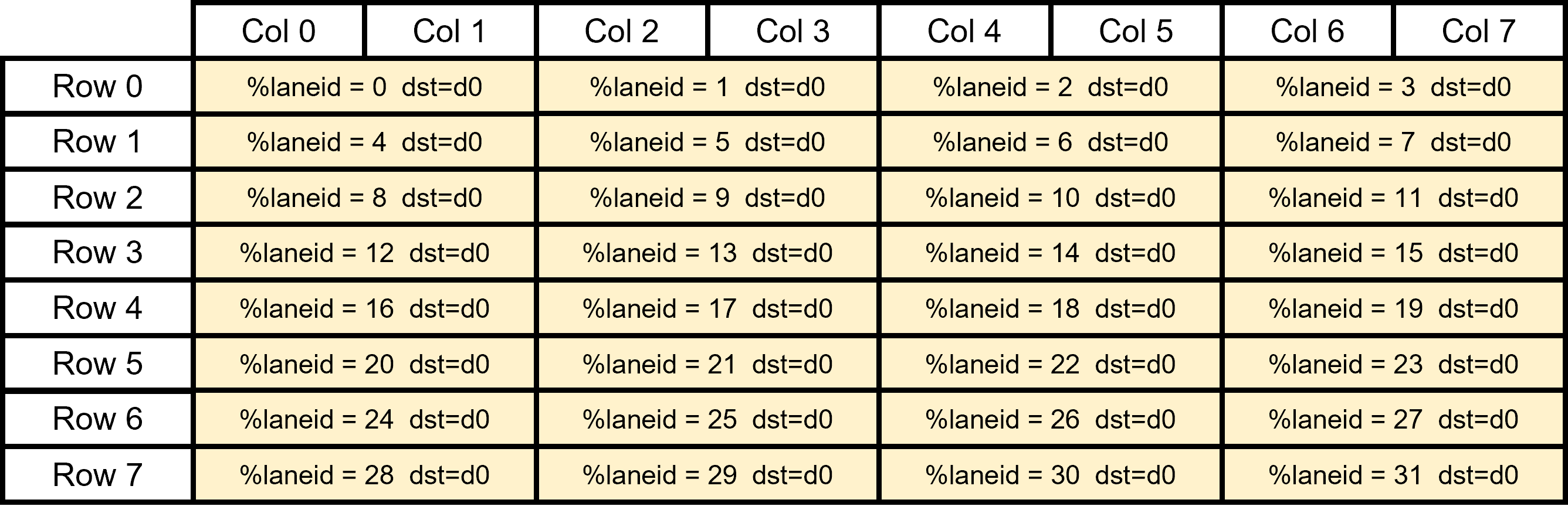}
    \caption{Data loading pattern of \textit{ldmatrix} instruction for a single $8\times8$ matrix. Two half-precision elements are loaded to the destination register \textit{d0} per each thread lane in a warp.}
    \label{fig:ldmatrix-pattern}
\end{figure}

Recent advancements in NVIDIA GPUs have showcased significant performance improvements in GEMM computation through the utilization of Tensor Cores. These Tensor Core-based GEMMs leverage warp-level PTX instructions, namely \textit{ldmatrix} and \textit{mma}. The \textit{ldmatrix} instruction efficiently loads multiple matrices across all threads within a warp from shared memory into designated registers. As illustrated in Figure~\ref{fig:ldmatrix-pattern}, this loading pattern assigns small fragments of a row to each thread, facilitating warp-level matrix multiply-accumulation using the subsequent \textit{mma} instruction.

The \textit{mma} instruction, following the \textit{ldmatrix} operation, executes the matrix multiply-accumulate operation at the warp level. This instruction performs the multiply-accumulate operation on matrices, requiring specific data patterns for each row of the multiplicands, as well as the accumulators. As previously described, loading matrices into each thread from shared memory is efficiently achieved using the \textit{ldmatrix} instruction.

Tensor Core-based GEMM computation entails repetitive calls to these instructions, relying on shared memory to rearrange input tensors to align with the data access pattern required by the \textit{mma} instruction. Compared to CUDA Core-based GEMM computation, Tensor Core-based approaches are renowned for achieving significantly higher throughput.

\subsection{Mixed precision GEMM kernel}

\begin{figure}
    \centering
    \includegraphics[width=1.0\columnwidth]{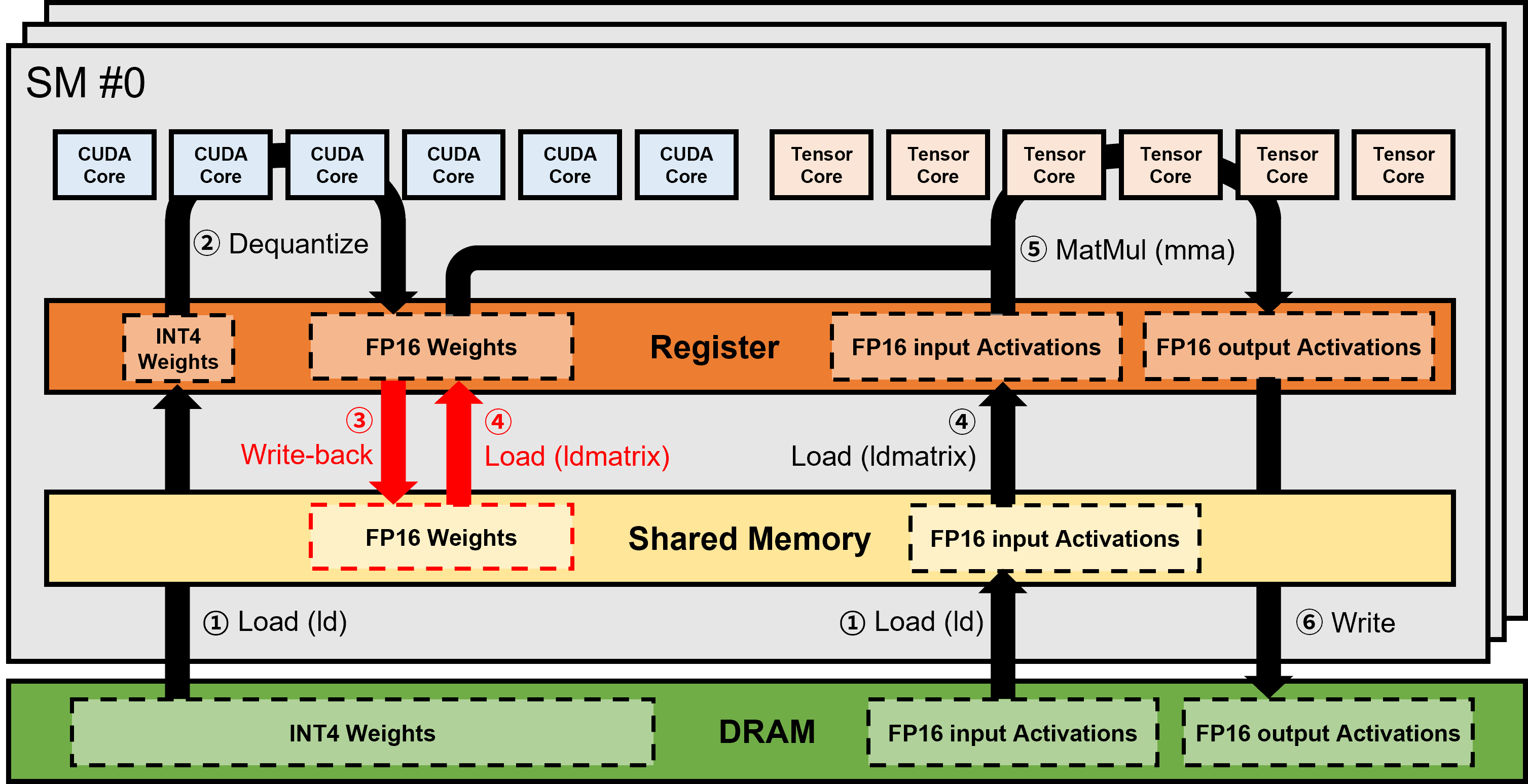}
    \caption{Computation overview of original kernel and QUICK. Compared to original kernel, QUICK bypasses 3) shared memory write-back and 4) \textit{ldmatrix} operation of dequantized weights by using interleaving data pattern.}
    \label{fig:compute-path}
\end{figure}

Mixed precision GEMM kernels find widespread application in the inference phase of weight-only quantized LLMs, owing to the inherent difference in bit precision between activation tensors and weight tensors.

When employing weight-only quantization, it becomes necessary to dequantize the quantized weights before executing the matrix multiplication operation within the GEMM kernel, as recent NVIDIA GPUs' Tensor Cores do not inherently support mixed-precision GEMMs. Consequently, numerous implementations of efficient mixed-precision GEMM kernels leveraging Tensor Cores adopt parallel dequantization of quantized weights.

Typically, these kernels adhere to a common workflow for weight dequantization, as depicted in Figure \ref{fig:compute-path}. They fetch quantized and packed weights from global memory to registers, dequantize weights using CUDA cores, and then write the dequantized weights back to shared memory for the following \textit{ldmatrix} instruction. The dequantization process employing CUDA cores involves bitwise AND operations to extract target sub-byte weights, bitwise SHIFT operations to rearrange bit positions, and parallel half-precision additions and multiplications to apply zero points and scales.

Parallel dequantization involves expanding quantized weights to larger bit sizes. For example, from 128-bit weight vectors consisting of 32 4-bit weights, dequantization produces 512-bit weight vectors containing 32 16-bit floating-point weights. This results in quantized weights that are four times larger compared to their full precision counterparts under same bandwidth requirement, increasing the burden of shared memory write-back. The augmented quantity of weights exacerbates shared memory bank conflicts during the write-back process of dequantized weights.

\begin{figure}
    \centering
    \includegraphics[width=1.0\columnwidth]{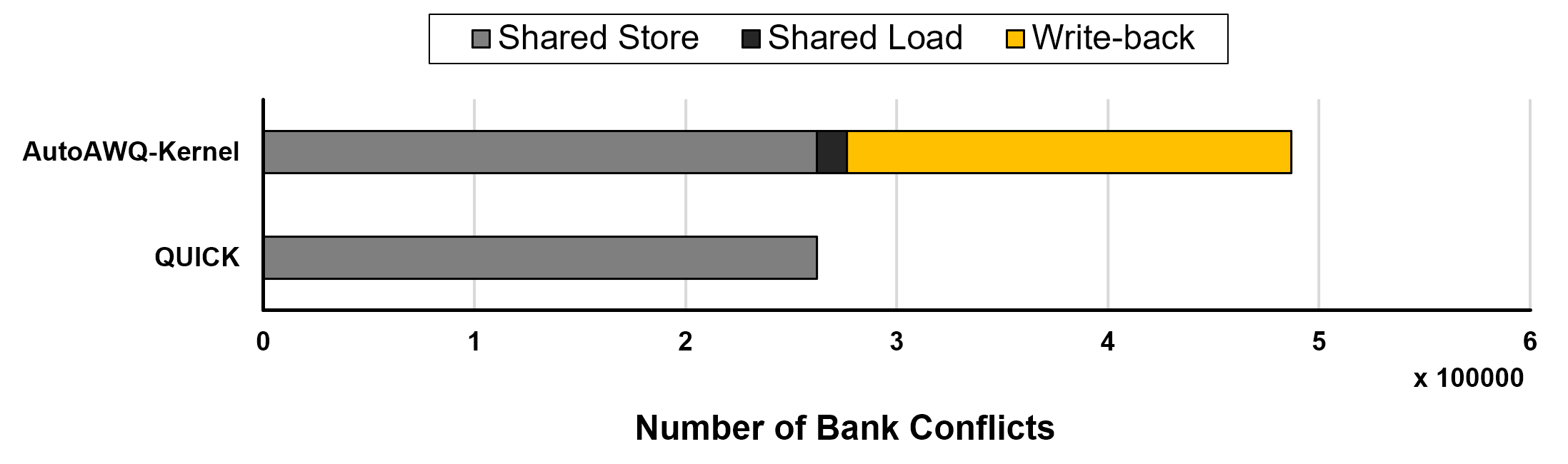}
    \caption{Number of bank conflicts from benchmark result using NVIDIA Nsight Compute \cite{nsight}. A matrix multiplication of shape $64 \times 8192 \times 8192 (M \times N \times K)$ was used as the workload.}
    \label{fig:bank-conflicts}
\end{figure}

Given that the \textit{ldmatrix} instruction requires the weight matrices to be fully visible on shared memory, this significantly harms the end-to-end latency of mixed-precision matrix multiplication. Benchmarks conducted on state-of-the-art mixed-precision GEMM kernels using NVIDIA's Nsight Compute \cite{nsight} indicate a notable prevalence of shared memory bank conflicts stemming from the write-back after dequantization, as depicted in Figure \ref{fig:bank-conflicts}. Consequently, mixed-precision GEMM kernels often struggle to achieve enhanced throughput compared to half-precision GEMM kernels, particularly with larger batch sizes.

\section{Avoiding Bank Conflict}
In this section, we propose QUICK, a novel way to remove the shared memory write-back bank conflicts of mixed precision matrix multiplication. To alleviate these conflicts effectively, our proposal involves reordering the quantized weight matrix offline to align with the load pattern required by the \textit{mma} instruction without the \textit{ldmatrix} instruction.

\subsection{Skipping Shared Memory Write-back During Mixed Precision GEMM}
As previously discussed, state-of-the-art mixed precision GEMM kernels rely on a specific sequence involving dequantization, shared memory write-back, \textit{ldmatrix}, and \textit{mma}. The \textit{ldmatrix} instruction is responsible for loading operands for the subsequent \textit{mma} instruction, adhering to a designated pattern among the threads within a warp. With this instruction, each thread in a warp loads fragments of a row, as depicted in Figure \ref{fig:ldmatrix-pattern}.

Using the \textit{ldmatrix} instruction to load GEMM operands to registers is a straightforward approach for floating-point GEMM kernels because transferring data from global memory to shared memory can be efficiently executed. From the Ampere architecture and beyond, asynchronous CUDA memory copy supports pipelining the \textit{mma} instruction with global memory load, thereby enhancing the performance of GEMM kernels. This enhancement occurs as the effective memory load overhead can be reduced to the copy from shared memory to registers.
However, in the case of mixed precision GEMM, there exists a noticeable overhead due to shared memory write-back. This is because the loaded quantized weights must be dequantized using CUDA cores, and the resulting dequantized weights in registers must then be written back to shared memory to serve as operands for the \textit{ldmatrix} instruction. This overhead is further exacerbated by numerous shared memory bank-conflict stalls, which ultimately degrade the throughput of kernels.

From the data loading pattern of the \textit{ldmatrix} instruction, we observe that this pattern can be pre-applied to the original data since the weight data remains static. Considering the static nature of weight matrices throughout deployment, it becomes feasible to bypass the \textit{ldmatrix} instruction for quantized weight matrices via suitable reordering. In this scenario, a direct load from global memory to registers proves sufficient to meet the data pattern requirements essential for the \textit{mma} operation. Consequently, we opt to rearrange the quantized weight matrices and bypass the \textit{ldmatrix} instruction prior to the \textit{mma} operation. Through the optimization of both the weight pattern and the associated computing kernel, we can successfully eliminate shared memory write-back bank conflicts, consequently improving the end-to-end latency of mixed precision GEMM. Importantly, since the total amount of quantized weights to be read from DRAM remains the same, the overall memory bandwidth requirement can be maintained at the same level.

\subsection{Interleaving Data Pattern}

\begin{figure}
    \centering
    \includegraphics[width=1.0\columnwidth]{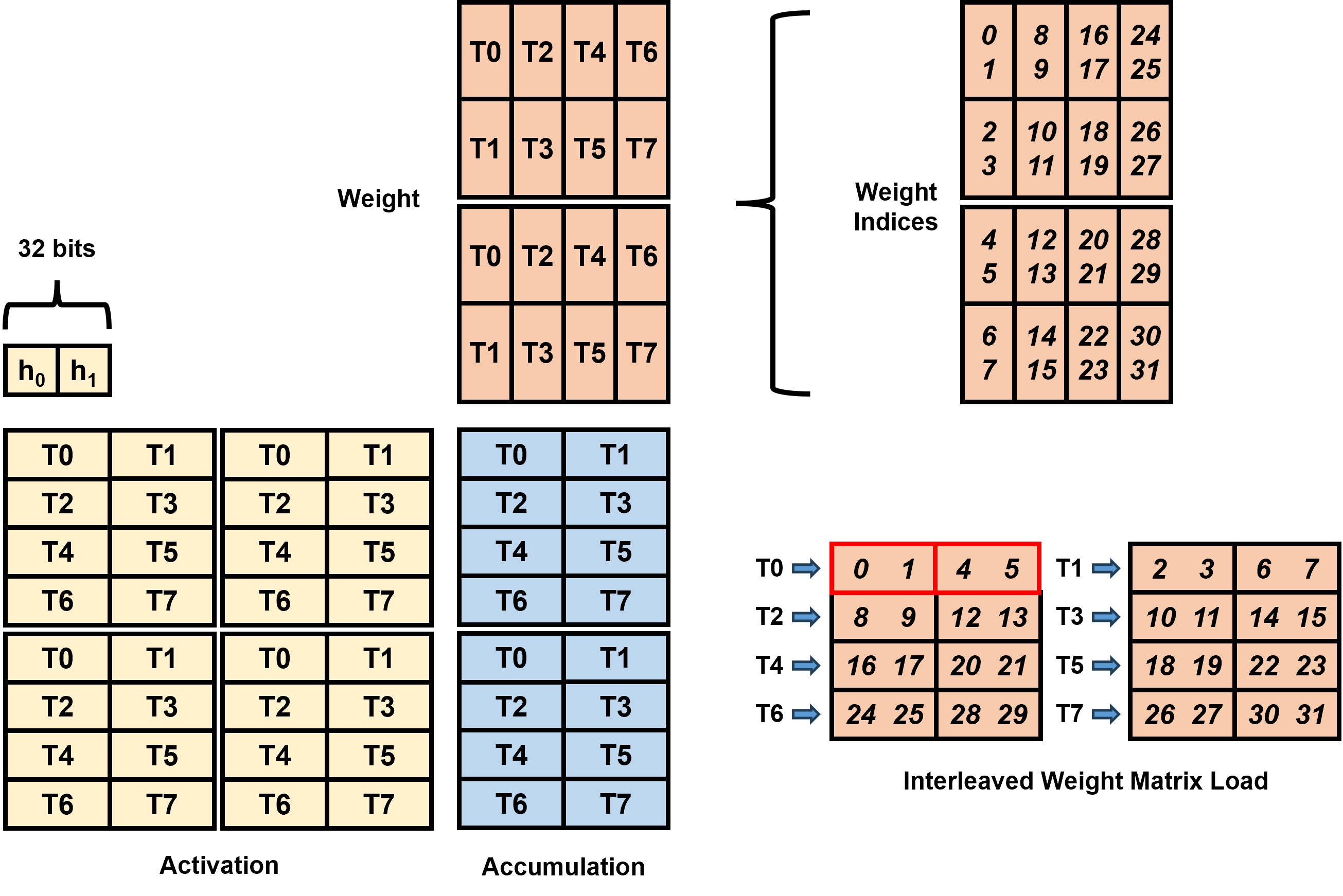}
    \caption{\textit{ldmatrix} instruction-aware weight interleaving to avoid shared memory conflicts. With interleaved weight matrix, direct load from DRAM to registers for each thread without \textit{ldmatrix} is possible. Note that the figure is illustrating a case of computing $8 \times 4 \times 8 (M \times N \times K)$ GEMM using 8 threads for simplicity. QUICK utilizes $16 \times 8 \times 16 (M \times N \times K)$ GEMM with 32 threads and its corresponding interleaving pattern.}
    \label{fig:ldmatrix-reorder}
\end{figure}

The interleaving pattern of the quantized weight matrices corresponds to the data loading pattern of the \textit{ldmatrix} instruction. To bypass the \textit{ldmatrix.sync.aligned.m8n8} instruction of the quantized weight matrices, we rearrange the weights following the data loading pattern of the instruction, as illustrated in Figure \ref{fig:ldmatrix-reorder}. Since the CUDA kernels of QUICK rely on the \textit{mma.m16n8k16} with half-precision, we further devise the reordering pattern to group quantized weights for two 8$\times$8 weight blocks. This rearrangement pattern enhances the memory locality of quantized weights and eliminates shared memory write-back bank conflicts.

\begin{figure}
    \centering
    \includegraphics[width=1.0\columnwidth]{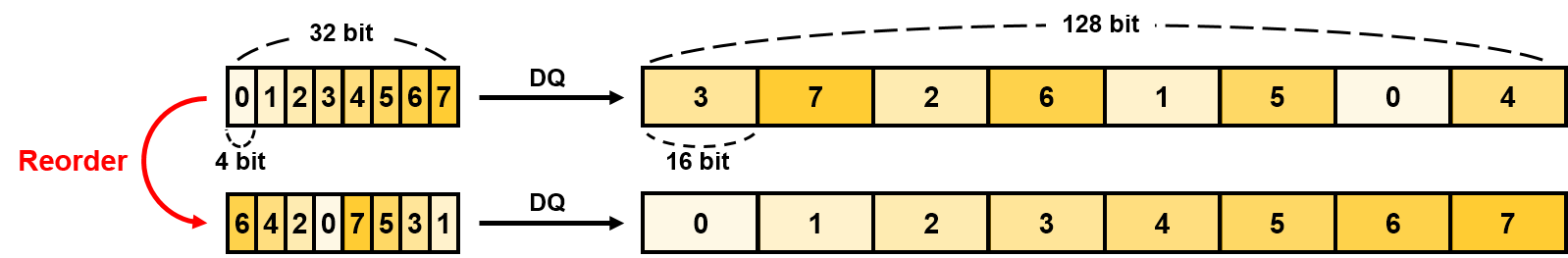}
    \caption{Parallel i4-f16 dequantization kernel-aware weight reordering.}
    \label{fig:deq-reorder}
\end{figure}

Moreover, QUICK implements an additional rearrangement of quantized weights based on the pattern of the dequantization kernel. QUICK utilizes a modified version of the parallel dequantization kernel from FasterTransformer \cite{nvidia2022fastertransformer}. The kernel introduces a simple interleaved pattern, as shown in Figure \ref{fig:deq-reorder}. To mitigate the overhead associated with rearranging the dequantized weights and further enhance data locality, an additional rearrangement pattern ensuring a sequential weight pattern after dequantization is applied.

\begin{figure}
    \centering
    \includegraphics[width=1.0\columnwidth]{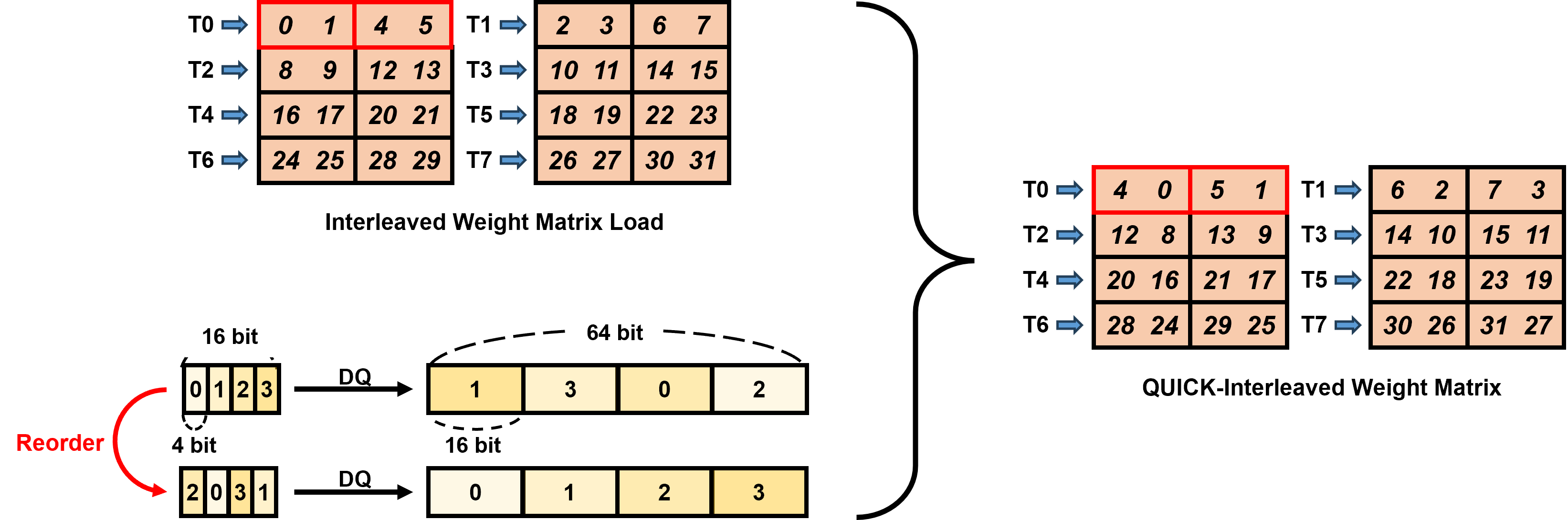}
    \caption{QUICK weight interleaving pattern. Note that the figure is illustrating a case of computing $8 \times 4 \times 8 (M \times N \times K)$ GEMM using 8 threads and 16-bit packed i4-f16 dequantization kernel for simplicity. QUICK utilizes $16 \times 8 \times 16 (M \times N \times K)$ GEMM with 32 threads and 32-bit packed i4-f16 dequantization kernel.}
    \label{fig:combined-reorder}
\end{figure}

Both weight rearrangement patterns avoid shared memory write-backs and ensure the sequential weight pattern after dequantization can be applied concurrently, as the patterns are independent. QUICK integrates both patterns as described in Figure \ref{fig:combined-reorder}, achieving optimal end-to-end latency while reducing shared memory bank conflicts and enhancing data locality.

\subsection{Tile Size Optimization}
Optimizing the number of active warps per multiprocessor plays an important role in improving the performance of computation kernels. Achieving higher number of active warps per multiprocessor can be beneficial as it facilitates the interleaving of warps and enables better latency hiding. Several factors, including the number of required registers and the size of shared memory, can limit the number of active warps per multiprocessor. In addition to improving throughput by eliminating shared memory write-back bank conflicts, QUICK leverages the reduced shared memory usage within the computation kernel to further enhance computational throughput.

Previous mixed precision GEMM kernels have utilized shared memory to store both activation and weight matrices, with benchmarks indicating that the shared memory size per warp exerts the greatest pressure on the number of active warps per multiprocessor. In contrast, QUICK avoids allocating shared memory for the weight matrices, thereby shifting the pressure from shared memory size to the number of required registers. Leveraging this opportunity, QUICK increases the tile size of mixed precision GEMM, further reducing DRAM accesses while maintaining similar theoretical multiprocessor occupancy. With increased number of activation values processed per computation tile, weight matrices need to be loaded less frequently from DRAM. This optimization results in a further increase in throughput for larger batch sizes, particularly those exceeding 32.

\section{Experimental Results}

In this section, we evaluate the performance improvement provided by QUICK in comparison to both the baseline fp16 kernel and AutoAWQ-Kernel. 
We first compare the efficiency of a single matrix multiplication, followed by the comparison of end-to-end token generation throughput across various LLMs. 
Furthermore, we also present the benchmark results showcasing the integration of QUICK with the vLLM~\cite{kwon2023efficient} framework.
Note that all experiments involving AutoAWQ-Kernel and QUICK are based on 4-bit weight-only quantization.

\subsection{Matrix Multiplication Performance}
\begin{figure}
    \centering
    \includegraphics[width=1.0\columnwidth]{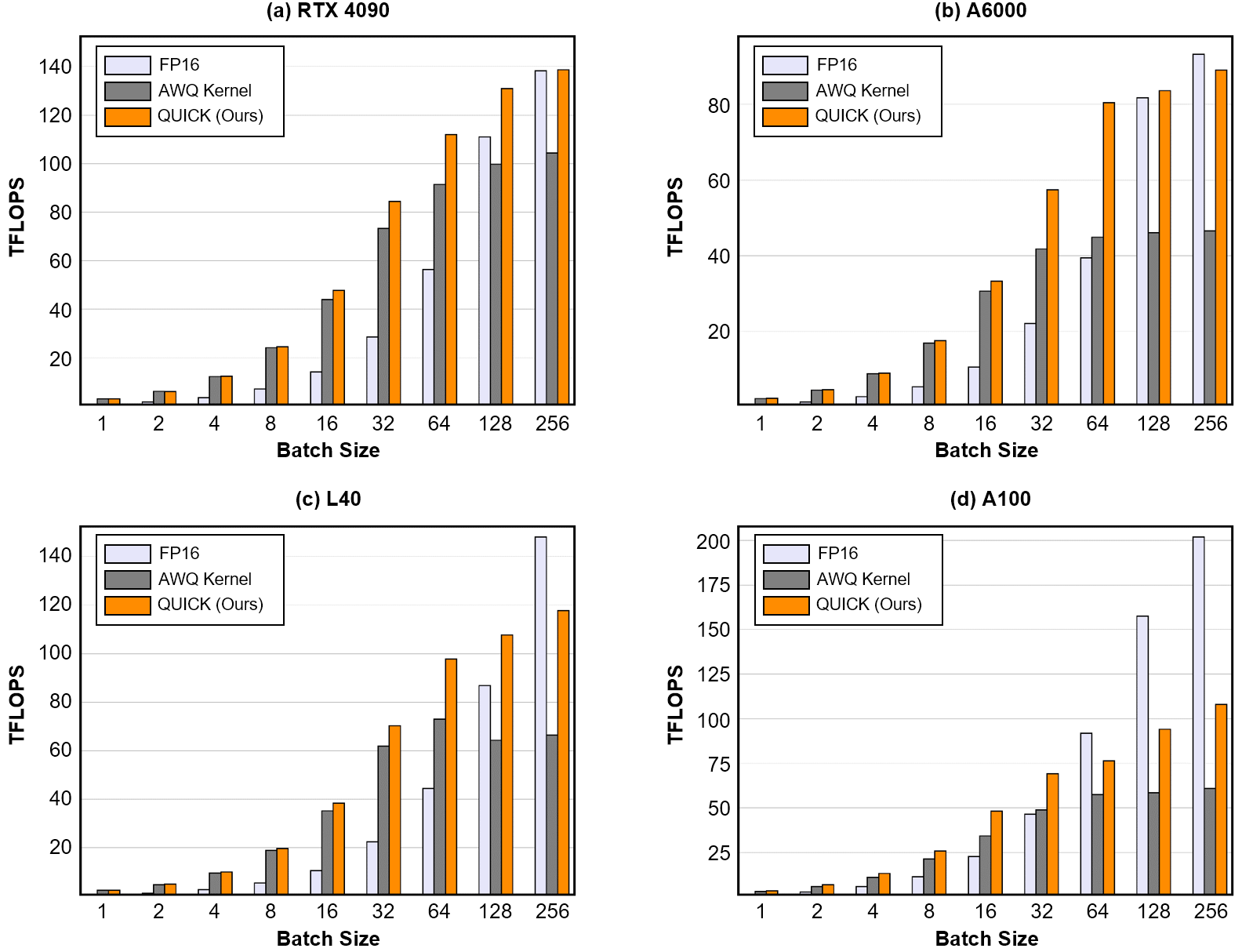}
    \caption{Benchmark results of matrix multiplication kernels on various GPUs. The shape of matrices is set to $\textit{batch size} \times 8192 \times 8192 (M \times N \times K)$.}
    \label{fig:kernel_benchmark}
\end{figure}

We initially evaluate the performance of QUICK with unit matrix multiplications, with the matrix multiplication dimensions set to $\textit{batch size} \times 8192 \times 8192 (M \times N \times K)$.
Figure \ref{fig:kernel_benchmark} illustrates the performance of different kernels in terms of Tera-operations per second (TOPS).
Notably, QUICK demonstrates superior performance compared to previous implementations such as AutoAWQ-Kernel \cite{hansen2023autoawqkernel}, particularly evident with larger batch sizes.
For instance, when the batch size is 256, QUICK demonstrates a speed improvement of 1.33$\sim$1.91 times compared to AutoAWQ-Kernel.

With larger batch sizes, the token generation process tends to become computation-bounded, making the overhead from the dequantization process more significant.
As a result, AutoAWQ-Kernel tends to show prominently degraded throughput compared to fp16 kernel when the batch size approaches 128.
On the other hand, QUICK, by reducing shared memory bank conflict problem, occasionally demonstrates faster speeds than the fp16 kernel, even with larger batch sizes like 128.

\subsection{End-to-end Throughput}

\begin{figure}
    \centering
    \includegraphics[width=0.9\columnwidth]{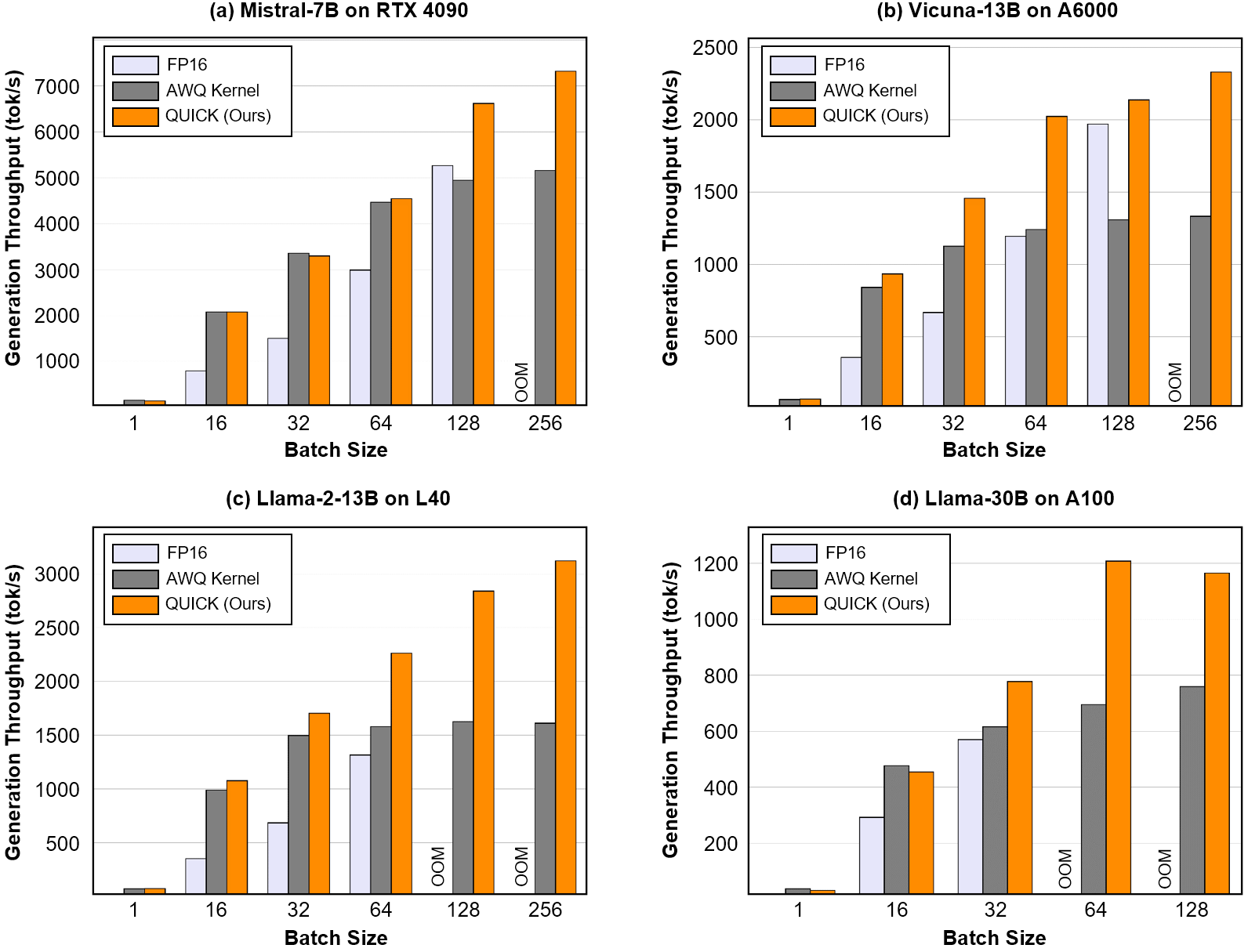}
    \caption{End-to-end token generation throughput benchmarks of (a) Mistral-7B \cite{jiang2023mistral} on RTX 4090, (b) Vicuna-13B \cite{vicuna2023} on RTX A6000, (c) LLaMA-2-13B \cite{roumeliotis2023llamav2} on L40, and (d) LLaMA-33B \cite{touvron2023llama} on A100.}
    \label{fig:e2e_benchmark}
\end{figure}

To illustrate the advantages of QUICK in the inference of quantized LLMs, we further evaluate the end-to-end token generation speed of various LLMs.
We conducted tests on four different models across four different GPUs: Mistral-7B \cite{jiang2023mistral} on RTX 4090, Vicuna-13B \cite{vicuna2023} on RTX A6000, LLaMA-2-13B \cite{roumeliotis2023llamav2} on L40, and LLaMA-33B \cite{touvron2023llama} on A100.
The token generation throughput at the decoding stage was measured in terms of tokens per second.

As the batch size increases, the memory required to store activations and the KV cache also increases, leading to Out-of-Memory (OOM) problem.
For example, when running Mistral-7B on an RTX 4090 GPU, it is impossible to run the fp16 model with batch size of 256 due to the OOM problem.
Applying weight-only quantization reduces the amount of memory used to store weights, thereby enabling usage of more memory for storing activations and the KV cache.
Consequently, larger batch inference becomes possible.
Even with the same RTX 4090 GPU, a 4-bit quantized Mistral-7B can be operated at a batch size of 256. Moreover, QUICK can achieve up to 1.94 times higher throughput compared to AutoAWQ-Kernel.
Similar to the Matrix Multiplication performance mentioned in the previous section, QUICK demonstrates superior performance over the fp16 case even at larger batch sizes.

\subsection{vLLM Throughput}
In this section, we present the throughput benchmark results of our initial version of vLLM \cite{kwon2023efficient} integrated with QUICK (Table \ref{tab:vllm_throughput}). Benchmarks were done using the throughput benchmark script and the recommended dataset within the vLLM \cite{kwon2023efficient} framework. Two models, Vicuna-13B \cite{vicuna2023} and Llama-2-70B \cite{roumeliotis2023llamav2}, were benchmarked to demonstrate scenarios where the full precision model could and could not be loaded onto the GPU device. vLLM with QUICK demonstrated a throughput gain of 27-29\% compared to the AWQ implementation in vLLM, and a 33\% throughput gain compared to the full precision model.

\begin{table}[ht]
    \caption{Throughput benchmark results of Vicuna-13B \cite{vicuna2023} and Llama-2-70B \cite{roumeliotis2023llamav2} models with vLLM \cite{kwon2023efficient} integrated with QUICK. Benchmarks were conducted on a machine equipped with an i9-13900K CPU, 128GB RAM, and an A6000 GPU.}
    \centering
    \begin{tabular}[t]{cccccc}
        \toprule
        \multirow{2}{*}{Model} & FP16 & AWQ & QUICK & Speedup & Speedup \\
        & (tokens/s) & (tokens/s) & (tokens/s) & (FP16) & (AWQ) \\
        \midrule
        Vicuna-13B & 985.2 & 1030.4 & 1308.6 & 33\% & 27\% \\
        Llama-2-70B & OOM & 224.3 & 290.2 & - & 29\% \\
        \bottomrule
    \end{tabular}
    \vspace{1.0pt}
    \label{tab:vllm_throughput}
\end{table}

\section{Limitation and Future Work}
While the proposed QUICK technique has demonstrated enhanced throughput at larger batch sizes, such as 128, enabling the utilization of weight-only quantization for larger batch sizes, it still falls short of the efficiency achieved in the fp16 case, particularly at even larger batch sizes (> 512).
Therefore, further research is needed to optimize the dequantization process further and enhance the efficiency of mixed precision GEMM kernels under such circumstances.

For instance, future works could focus on exploring methods to leverage the unused shared memory budget resulting from the direct dequantization of quantized weights at registers. Additional software optimizations, such as automated split-k parameter optimization, could be explored further to ensure optimal throughput considering the model, generation configuration, and the GPU device.

\section{Conclusion}
In this work, we introduce QUICK, a suite of optimized CUDA kernels designed for efficient execution of mixed precision GEMM operations. 
Previous implementations exhibited advantages only for small batch sizes due to shared memory bank conflict problem. 
QUICK, however, overcomes this limitation by employing an interleaving data pattern, which enables superior throughput over fp16 kernels even for larger batch sizes.
Furthermore, QUICK has demonstrated enhanced end-to-end token generation throughput in various LLM inference frameworks, including AutoAWQ and vLLM.

\def\bibfont{\small}

\bibliographystyle{plain} 
\bibliography{ref} 

\end{document}